\documentclass{article}

     \usepackage[final]{neurips_2019}

\usepackage[utf8]{inputenc} 
\usepackage[T1]{fontenc}    
\usepackage{hyperref}       
\usepackage{url}            
\usepackage{booktabs}       
\usepackage{amsfonts}       
\usepackage{nicefrac}       
\usepackage{microtype}      
\usepackage{graphicx,subcaption}
\usepackage{amsthm}
\usepackage{placeins}

\usepackage{amsmath,amsfonts,bm}

\def\eqref#1{equation~\ref{#1}}

\def\1{\bm{1}}

\DeclareMathAlphabet{\mathsfit}{\encodingdefault}{\sfdefault}{m}{sl}
\SetMathAlphabet{\mathsfit}{bold}{\encodingdefault}{\sfdefault}{bx}{n}

\newcommand{\E}{\mathbb{E}}

\DeclareMathOperator*{\argmax}{arg\,max}
\DeclareMathOperator*{\argmin}{arg\,min}

\usepackage[ruled,vlined]{algorithm2e}
\theoremstyle{definition}
\newtheorem{definition}{Definition}[section]

\theoremstyle{remark}

\usepackage[ruled,vlined]{algorithm2e}
\SetKwInput{KwInit}{Init}
\SetKwInput{KwGiven}{Given}
\SetKwInput{KwReturn}{Return}
\SetKwComment{Comment}{start}{end}
\usepackage{booktabs}
\usepackage{multirow}
\newcommand{\cmmnt}[1]{\ignorespaces}

\usepackage{mdwlist}
\usepackage{authblk}
\newcommand*{\affaddr}[1]{#1} 

\newcommand*{\email}[1]{\texttt{#1}}
\newcommand\blfootnote[1]{%
  \begingroup
  \renewcommand\thefootnote{}\footnote{#1}%
  \addtocounter{footnote}{-1}%
  \endgroup
}

\usepackage{hyperref}
\usepackage{url}
\title{Fairness With Minimal Harm:\\ A Pareto-Optimal Approach For Healthcare}

\author{%
Natalia Martinez$^\dagger$, Martin Bertran$^\dagger$, Guillermo Sapiro\\
\affaddr{Department of Electrical and Computer Engineering}\\
\affaddr{Duke University}\\
\email{\{natalia.martinez,martin.bertran,guillermo.sapiro\}@duke.edu}\\
}

\usepackage{fancyhdr}
\begin{document}

\maketitle

\begin{abstract}
Common\blfootnote{$\dagger$ Equal contributions} fairness definitions in machine learning focus on balancing notions of disparity and utility. In this work, we study fairness in the context of risk disparity among sub-populations. We are interested in learning models that minimize performance discrepancies across sensitive groups without causing unnecessary harm. This is relevant to high-stakes domains such as healthcare, where non-maleficence is a core principle. We formalize this objective using Pareto frontiers, and provide analysis, based on recent works in fairness, to exemplify scenarios were perfect fairness might not be feasible without doing unnecessary harm. We present a methodology for training neural networks that achieve our goal by dynamically re-balancing subgroups risks. We argue that even in domains where fairness at cost is required, finding a non-unnecessary-harm fairness model is the optimal initial step. We demonstrate this methodology on real case-studies of predicting ICU patient mortality, and classifying skin lesions from dermatoscopic images.
\end{abstract}

\section{Introduction}
Machine learning algorithms play an important role in decision making in society. When these algorithms are used to make high-impact healthcare decisions such as deciding whether or not to classify a skin lesion as benign or not, or predicting mortality for intensive care unit patients, it is paramount to guarantee that these decisions are accurate and unbiased with respect to sensitive attributes such as gender or ethnicity. A model that is trained naively may not have these properties by default (\cite{barocas2016big}). 
It is desirable in these critical applications to impose some fairness criteria. There are several lines of work in fairness in machine learning, such as Demographic Parity (\cite{louizos2015variational,zemel2013learning,feldman2015certifying}), Equality of Odds, Equality of Opportunity (\cite{hardt2016equality, woodworth2017learning}), or Calibration (\cite{pleiss2017fairness}). These notions of fairness are appropriate in many scenarios, but in domains where quality of service is paramount, such as healthcare, we argue that it is necessary to strive for models that are as close to fair as possible without introducing unnecessary harm to any subgroup (\cite{ustun2019fairness}). 

In this work, we measure discrimination (unfairness) in terms of difference in predictive risks across sub-populations defined by our sensitive attributes. This concept has been explored in other works such as \cite{calders2010three,dwork2012fairness,feldman2015certifying,chen2018my,ustun2019fairness}. We examine the subset of models from our hypothesis class that have the best trade-offs between sub-population risks, and select from this set the one with the smallest risk disparity gap. \cmmnt{The goal is to select an algorithm from our hypothesis class that has the smallest risk disparity gap between subgroups, while at the same time being as close to optimal performance in every subgroup as possible.} This is in contrast to common post-hoc correction methods like the ones proposed in \cite{hardt2016equality,woodworth2017learning} where randomness is potentially added to the decisions of all sub-populations. While this type of approach diminishes the accuracy-disparity gap, it does so by potentially introducing randomness into the final decision (i.e., with some probability, disregard classifier output and produce arbitrary decision based solely on sensitive label), which leads to performance degradation in most common risk metrics. Since our proposed methodology does not require test-time access to sensitive attributes, and can be applied to any standard classification or regression task, it can also be used to reduce risk disparity between outcomes, acting as an adaptive risk equalization loss compatible with unbalanced classification scenarios.

\paragraph{Main Contributions} We formalize the notion of no-unnecessary-harm fairness using Pareto optimality, a state of resource allocations from which it is impossible to reallocate without making one subgroup worse. We show that finding a Pareto-fair classifier is equivalent to finding a model in our hypothesis class that is both Pareto optimal with respect to the sub-population risks (no unnecessary harm) and minimizes risk disparity. This notion is already amenable to non-binary sensitive attributes. We analyze Pareto fairness on an illustrative example, and compare it to alternative approaches. We provide an algorithm that promotes fair solutions belonging to the Pareto front; this algorithm can be applied to any standard classifier or regression task. Finally, we show how our methodology performs on real tasks such as predicting ICU mortality rates in the MIMIC-III dataset from hospital notes \cite{johnson2016mimic}, and classifying skin lesions in the HAM10000 dataset \cite{tschandl2018ham10000}.

\section{Problem Statement}

Consider we have access to a dataset $\mathcal{D} = \{(x_i,y_i,a_i)\}_{i=1}^n$ containing  $n$ independent triplet samples drawn from a joint distribution $(x_i,y_i,a_i) \sim P(X,Y,A)$ where $x_i \in \mathcal{X}$ are our input features (e.g., images, tabular data, etc.), $y_i \in \mathcal{Y}$ is our target variable, and $a_i \in \mathcal{A}$ indicates group membership or sensitive status (e.g., ethnicity, gender); our input features $X$ may or may not explicitly contain $A$. 

Let $h \in \mathcal{H}$ be a classifier from our hypothesis class $\mathcal{H}$ trained to infer $y$ from $x$, $h:\mathcal{X} \rightarrow \mathcal{Y}$; and a loss function $\ell: \mathcal{Y}\times \mathcal{Y}  \rightarrow \mathbb{R}$. We define the class-specific risk of classifier $h$ on subgroup $a$ as $R_a(h) = \E_{X,Y|A=a}[\ell(Y,h(X))]$. The risk discrimination gap between two subgroups $a,a' \in \mathcal{A}$ is measured as  $\Gamma_{a,a'}(h) = |R_a(h)- R_{a'}(h)|$, and we define the pairwise discrimination gap as $\Vec{\Gamma}_\mathcal{A}(h) = \{\Gamma_{a,a'}(h)\}_{a,a' \in \mathcal{A}}$. Our goal is to obtain a classifier $h \in \mathcal{H}$ that minimizes this gap without causing unnecessary harm to any particular group. To formalize this notion, we define:


\begin{definition}{Pareto front:}
The set of Pareto front classifiers is defined as $\mathcal{P}(\mathcal{H},\mathcal{A}) = \{ h \in \mathcal{H}: \not\exists h' \in \mathcal{H}, \not\exists a' \in \mathcal{A}: [R_{a'}(h') < R_{a'}(h)] \land [R_a(h') \le R_a(h)], \forall a \in \mathcal{A}   \} $
\label{def:paretofront}
\end{definition}

\begin{definition}{Pareto-fair classifier and Pareto-fair vector:}
A classifier $h^*$ is an optimal no-harm classifier if it minimizes the discrimination gap among all Pareto front classifiers, $h^* = \argmin\limits_{h \in \mathcal{P}(\mathcal{H},\mathcal{A})} || \Vec{\Gamma}_\mathcal{A}(h) ||_{\infty} $. The Pareto-fair vector is defined as $\bm{r}^* \in \mathbb{R}^{|\mathcal{A}|} :$  $\bm{r}^* = \{R_a(h^*)\}_{a\in \mathcal{A}}$.
\label{def:noharm}
\end{definition}

\begin{figure}[ht]
    \centering
    \includegraphics[width=.9\linewidth]{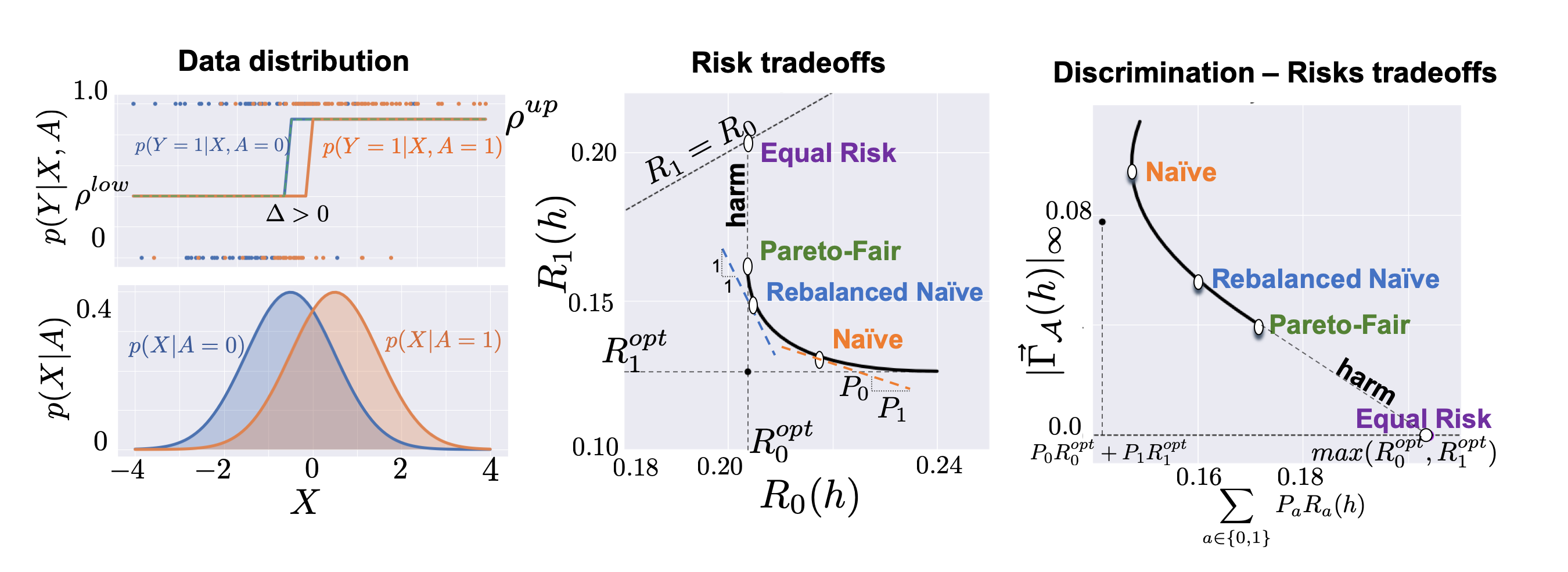}
    \caption{Example of binary target ($y$) and sensitive ($a$) variable. Bottom left figure shows conditional distributions of observation variable ($x$) $p(x|a)$, while top left shows distribution of $y$ as a function of $x,a$, $p(y|x,a)$, these distributions are simple piece-wise constants with two levels $\rho^{low}$ and $\rho^{high}$, $\Delta$ represents the gap between level transitions for each group. Middle image shows the Pareto front between group risks, since noise levels are not the same across subgroups, perfect fairness is not attainable. Equality of Risk requires pure degradation of service for group $a=1$. Both Naive and Re-balanced Naive classifiers do not attain the best possible no-harm classifier. Pareto-fair point is shown in green. Rightmost figure shows the trade-offs attainable between discrimination and mean risks, an equivalent problem to the risk trade-off figure.}

    \label{fig:Gaussians}
\end{figure}

The Pareto front defines the best achievable trade-offs between population risks $R_a(h)$, while the Pareto-fair classifier gives the trade-off with least disparity. Building on analysis done by \cite{chen2018my,domingos2000unified}, it is possible to decompose the risk in bias, variance and noise for some given loss functions. The noise represents the smallest achievable risk for infinitely large datasets. If it differs between sensitive groups, zero-discrimination (perfect fairness) can be only achieved by introducing bias or variance, hence doing harm. Figure \ref{fig:Gaussians} shows a scenario where the Pareto-front does not intersect the equality of risk line for the case of two sensitive groups $a \in \{0,1\}$ and a binary output variable $y \in \{0,1\}$. Here the level of noise between subgroups differs, and the Pareto-fair vector $\bm{r}^*$ would not be achieved by either a naive classifier (minimizes expected global risk), or a classifier where low-occurrence subgroups are over-sampled (re-balanced naive classifier). 

\section{Methods}
Any loss function that is monotonically decreasing with $R_{a}(h), \forall a$ is minimized by a classifier in the Pareto front. Since we want to minimize the discrimination gap, we will build an adaptive loss function that is shares this property with the following form:
\begin{equation}
\begin{array}{cc}
     \phi(h;\bm{\mu},c) &= \sum\limits_{a\in\mathcal{A}} R_{a}(h) +\mu_a (R_{a}(h)-c)^{+2},
\end{array}
\label{eq:adaptiveloss}
\end{equation}
with $c < \min_{a}R_{a}(h)$ and  $\bm{\mu} = \{ \mu_a\}_{a\in \mathcal{A}}$. It can be shown that for convex Pareto sets (with respect to risk vectors $\bm{r} = \{R_{a}(h)\}_{a \in \mathcal{A}}$), there exist a set of $\bm{\mu}^*$ such that the Pareto-fair vector $\bm{r}^*$ is a unique solution. In Algorithm \ref{algo:MainAlgo} we jointly search for  $\bm{\mu}^*$ ,$\bm{r}^*$, and $h^*$.

\begin{algorithm}[H]
\DontPrintSemicolon
\SetNoFillComment
\footnotesize
\SetAlgoLined
\KwGiven{$h_\theta, \mathcal{D},  \gamma>0, \xi\in(0,1), \zeta\in(0,1), \text{lr}, $}
$\bm{\mu} \leftarrow \bm{0},\; \bm{\mu}^* \leftarrow \bm{0},\;,c\leftarrow 0,\; h^* \leftarrow h_\theta $\;
\While{ Improving}{
    $h_\theta \leftarrow \text{SGDwithEarlyStopping}(h_\theta,\phi, \bm{\mu},c, \mathcal{D})  $\tcp*[l]{Optimize adaptive loss with fixed $\bm{\mu}$ on dataset $\mathcal{D}$ }
    $r^{\text{val}} \leftarrow \text{EvaluateRisk}(h_\theta, \mathcal{D}^{\text{val}})$\;
  \uIf{$ || \Vec{\Gamma}_\mathcal{A}(h) ||_{\infty} < \Gamma^*$ and $\bm{r}^{\text{Val}}$ is not dominated by previous validation risks }{
 $h^* \leftarrow h_\theta,\; \Gamma^* \leftarrow || \Vec{\Gamma}_\mathcal{A}(h) ||_{\infty},\; c_{\text{old}} \leftarrow c,\; c \leftarrow \min_a \frac{r_{a}^{\text{Val}}}{k}$\; 
 $ \bm{\mu}^* \leftarrow \bm{\mu}\cdot \frac{(\bm{r}^{\text{Val}}-c_{\text{old}})^+}{(\bm{r}^{\text{Val}}-c)^+},\; a' \leftarrow \argmax_a r_{a}^{\text{Val}}$}
 \Else{lr $\leftarrow$ $\zeta$ lr, $\bm{\mu} \leftarrow \bm{\mu}^*,\; \gamma \leftarrow \gamma \xi,\; h_\theta \leftarrow h^*$}
 $\mu_{a'}\leftarrow (1+\gamma) \mu_{a'}$
 }
 \KwReturn{$h^*$ \tcp*[1]{Exit loop due to excessive iterations or no improvement in fairness}}
 \caption{ParetoFairOptimization}
 \label{algo:MainAlgo}
\end{algorithm}

\section{Experiments and Results}
We evaluate our method on mortality prediction and skin lesion classification and show empirically how we reduce accuracy and risk disparity without unnecessary harm. We compare it against a naive classifier, class-rebalancing (equally sampled sub-groups), the post-processing framework presented in \cite{hardt2016equality}, and the Disparate Mistreatment framework of \cite{zafar2017parity}. 

\subsection{Predicting Mortality in Intensive Care Patients}

We analyze clinical notes from  adult ICU patients at the Beth Israel Deaconess Medical Center (MIMIC-III dataset) \cite{johnson2016mimic} to predict patient mortality. We follow the pre-processing methodology outlined in \cite{chen2018my} and use tf-idf statistics on the $10,000$ most frequent words in clinical notes as input features. Fairness is measured with respect to age (under/over 55 years old), ethnicity, and outcome. We used a fully connected neural network with two 2048-unit hidden layers trained with Brier score (BS) loss. Table \ref{table:MIMICResults} shows accuracy and BS of all tested methodologies. We observe from columns (PF BS) and (PF Acc) that our model has the best accuracy and BS discrepancies. We can reduce accuracy disparities further by applying Hardt post processing (HPF Acc), at the cost of reducing overall performance. Note that, as expected, it is better to apply this post processing on our method than on the rebalanced naive classifier (HPF Acc vs HRen Acc). This illustrates the goal of the proposed Pareto-fair paradigm: develop the fairest algorithm with no unnecessary harm, and if (e.g., due to policy) the resulting fairness level needs to be improved, use other post-processing techniques on the Pareto-fair classifier such as \cite{hardt2016equality}.

\begin{table}[ht]
\scriptsize
\centering
\begin{tabular}{lc|cccc|cc|cccc}
\toprule
 Out/Age/Race & Ratio  & Na Acc & ReN Acc &  Zafar Acc & PF Acc & HReN Acc & HPF Acc & ReN BS & PF BS \\ 
\midrule
 A/A/NW &5.7\%&  99.1$\pm$0.4\% &  86.3$\pm$1.5\% &  93.0$\pm$1.4\% &  83.4$\pm$2.6\% &76.3$\pm$1.9\% &
71.6$\pm$3.2\% & 0.2 $\pm$ 0.02 & 0.25 $\pm$ 0.03 
 \\
 A/A/W &13.3\% & 98.8$\pm$0.5\% &  86.3$\pm$1.1\%&  90.0$\pm$1.3\% &  83.2$\pm$1.5\% & 76.7$\pm$1.6\% &
71.8$\pm$1.9\%  & 0.2 $\pm$ 0.01 & 0.25 $\pm$ 0.02  \\
 A/S/NW &12.9\%  &  97.5$\pm$0.6\% &  76.5$\pm$1.7\%&  81.8$\pm$1.7\%  & 71.4$\pm$3.0\% & 76.4$\pm$2.2\% &
 71.3$\pm$3.1\%  & 0.31 $\pm$ 0.02 & 0.36 $\pm$ 0.03  \\
 A/S/W &56.7\%  & 97.9$\pm$0.3\% &  79.0$\pm$0.6\%  &  77.4$\pm$0.7\%& 74.6$\pm$1.6\% & 76.2$\pm$2.2\% &
72.1$\pm$1.2\%  & 0.28 $\pm$ 0.01 & 0.34 $\pm$ 0.02  \\
 D/A/NW &0.4\%  & 23.4$\pm$9.4\% &  76.1$\pm$8.5\% &  47.7$\pm$9.6\% &  78.6$\pm$6.1\% & 66.6$\pm$9.9\% & 
74.1$\pm$9.2\% & 0.36 $\pm$ 0.06 & 0.34 $\pm$ 0.04 \\
 D/A/W &0.9\%  & 32.6$\pm$3.7\% &  80.1$\pm$3.3\%&  60.5$\pm$6.3\%  & 83.3$\pm$4.2\% & 66.4$\pm$2.4\% & 
73.6$\pm$4.3\% & 0.29 $\pm$ 0.04 & 0.28 $\pm$ 0.03   \\
 D/S/NW &1.8\%  & 21.4$\pm$2.2\% &  66.9$\pm$2.4\% &  48.2$\pm$2.0\% &   73.3$\pm$2.9\% & 64.8$\pm$2.1\% &
71.2$\pm$2.9\%  & 0.42 $\pm$ 0.02 & 0.37 $\pm$ 0.03 \\
 D/S/W &8.3\%  & 23.4$\pm$2.2\% &  67.4$\pm$1.9\% &  57.1$\pm$2.2\% &  72.5$\pm$3.6\% &66.2$\pm$2.9\% & 
72.4$\pm$3.5\%  & 0.42 $\pm$ 0.02 & 0.37 $\pm$ 0.04 \\
 \textbf{Sample Mean} & - &  \textbf{89.5$\pm$0.2\%} &  78.9$\pm$0.7\% &  78.0$\pm$0.7\% &  75.7$\pm$1.1\% &     \textbf{75.1$\pm$1.8\%} & 
 71.9$\pm$1.2\%  & \textbf{0.29 $\pm$ 0.01} & 0.33 $\pm$ 0.01 \\
 \textbf{Group Mean} & 12.5\%  &  61.8$\pm$1.5\% &  77.3$\pm$1.3\% &  69.4$\pm$1.3\% & \textbf{77.5$\pm$0.7\%}  &  71.2$\pm$1.1\% &
\textbf{72.3$\pm$1.0\%} & \textbf{0.31 $\pm$ 0.01} & 0.32 $\pm$ 0.01 \\
 \textbf{Discrepancy} & 56.3\% &  81.1$\pm$2.5\% &  22.5$\pm$2.5\% &  49.3$\pm$4.4\% & \textbf{17.1$\pm$2.6\%} &18.6$\pm$3.2\% &
\textbf{13.1$\pm$5.6\%}  & 0.24 $\pm$ 0.03 & \textbf{0.16 $\pm$ 0.03} \\
\bottomrule
\end{tabular}
\caption{\footnotesize Result summary table on MIMIC dataset.  Sensitive attributes are the combination of Outcome (Alive or Deceased), Age (Adult or Senior) and Race (White or Non-White). Target attribute is Outcome. We report Brier scores (BS) and Accuracies (Acc) for Naive (Na), Rebalanced Naive (ReN), Zafar (Zafar), and Pareto-Fair (PF) classifiers; the prefix (H) indicates that Hardt post-processing was applied to control for Age and Race. Zafar was applied on top of the embeddings of the Naive classifier, as that produced the best results.}
\label{table:MIMICResults}
\end{table}

\subsubsection{Skin Lesion Classification}

The HAM10000 dataset \cite{tschandl2018ham10000} collects over $10,000$ dermatoscopic images of skin lesions over a diverse population, lesions are classified in 7 categories. We used a pretrained DenseNet121 \citet{huang2017densely} as our base classifier, and measured fairness with respect to diagnosis class, casting balanced risk minimization as a particular use-case for Pareto fairness. Table \ref{tab:HAMResults} shows our empirical results.

\begin{table}[!htb]
\centering
\scriptsize
\begin{tabular}{lc|ccc|ccc}

\toprule
        Groups &  Ratio & PF Acc &   ReN Acc &    Na Acc & PF BS &   ReN BS &    Na BS \\
\midrule
     akiec&   2.5\% &    51.9\% &  55.6\% &   3.7\% &    0.741 &  0.671 &  1.289  \\
     bcc&   2.7\% &    56.7\% &  76.7\% &  56.7\%&    0.549 &  0.341 &  0.613 \\
     bkl&   7.2\% &    59.5\% &  58.2\% &  36.7\%  &     0.62 &  0.606 &  0.931 \\
     df&   0.5\% &    66.7\% &  33.3\% &   0.0\% &    0.536 &  0.898 &  1.721 \\
     nv&  81.0\% &    83.5\% &  90.9\% &  96.0\%&    0.241 &  0.128 &  0.054 \\
     vasc&   1.3\% &    71.4\% &  85.7\% &   0.0\%&     0.36 &  0.246 &  1.604 \\
     mel&   4.8\% &    53.8\% &  48.1\% &  32.7\%&    0.586 &  0.657 &  0.941 \\
 Sample Mean &   - &    78.6\% &  \textbf{84.8}\% &  83.6\%&    0.308 &  \textbf{0.213} &  0.234 \\
  Group Mean &  14.3\% &    63.4\% &  \textbf{64.1\%} &  32.3\%  &    0.519 &  \textbf{0.507} &  1.022 \\
        Discrepancy &  80.4\% &    \textbf{31.7}\% &  57.5\% &  96.0\%  &    \textbf{0.501} &  0.769 &  1.667 \\
\bottomrule
\end{tabular}



 \caption{\footnotesize{} Group-specific risks and accuracies for Naive (Na), Rebalanced Naive (ReN), and Pareto-Fair (PF) classifiers. Lesion Type was chosen as both target and sensitive variable, they are classified as Actinic keratoses and intraepithelial carcinoma (akiec), basal cell carcinoma (bcc), benign keratosis-like lesions (bkl), dermatofibroma (df), melanoma (mel), melanocytic nevi (nv) and vascular lesions (vasc). The Pareto-fair classifier exhibits the lowest discrepancy on Acc and BS while still being Pareto-optimal.}
\label{tab:HAMResults}
\end{table}

\section{Discussion}

Here we explore the problem of reducing the risk disparity gaps in the most ethical way possible (i.e., minimizing unnecessary harm). We provide an algorithm that can be used with any standard classification or regression loss to bridge risk disparity gaps without introducing unnecessary harm. We show its performance on two real-world case studies; we take advantage of the fact that our method does not require test-time access to sensitive attributes to frame balanced classification as a fairness problem. In future work, we wish to analyze if we can automatically identify high-risk sub-populations as part of the learning process and attack risk disparities as they arise, rather than relying on preexisting notions of disadvantaged groups or populations. We believe that no-unnecessary-harm notions of fairness are of great interest for several applications, especially so on domains such as healthcare.

\bibliography{cites.bib}
\bibliographystyle{plainnat}

\end{document}